\pdfoutput=1

\documentclass[11pt]{article}

\usepackage{emnlp2021}
\usepackage{times}
\usepackage{latexsym}

\usepackage[T1]{fontenc}

\usepackage[utf8]{inputenc}

\usepackage{microtype}

\usepackage{amsmath}
\usepackage{hyperref}
\usepackage{url}
\usepackage{mathrsfs}
\usepackage{multirow}
\usepackage{booktabs}
\usepackage{graphicx}
\usepackage{subcaption}
\usepackage{cleveref}
\usepackage{CJK}
\usepackage{CJKspace}
\usepackage{fdsymbol}
\usepackage{enumitem} 

\crefname{equation}{Eq.}{Eq.}
\crefname{section}{Section}{Sections}
\crefname{subsection}{Section}{Sections}
\crefname{subsubsection}{Section}{Sections}
\crefname{figure}{Figure}{Figures}
\crefname{table}{Table}{Tables}
\crefname{subfigure}{Figure}{Figures}
\crefname{algocf}{Algorithm}{Algorithms}
\usepackage{pifont}
\title{Weakly Supervised Contrastive Learning for Chest X-Ray Report Generation}

\author{An Yan, Zexue He, Xing Lu, Jiang Du, Eric Chang, \\ 
\textbf{Amilcare Gentili, Julian McAuley, Chun-Nan Hsu} \\
University of California, San Diego\\
\texttt{\{ayan,zehe,xil135,jiangdu,e8chang\}@ucsd.edu} \\ 
\texttt{\{agentili,jmcauley,chunnan\}@ucsd.edu} \\
}

\begin{document}
\maketitle
\begin{abstract}
Radiology report generation aims at generating descriptive text from radiology images automatically, which may present an opportunity to improve radiology reporting and interpretation. A typical setting consists of training encoder-decoder models on image-report pairs with a cross entropy loss, which struggles to generate informative sentences for clinical diagnoses since normal findings dominate the datasets. 
To tackle this challenge and encourage more clinically-accurate text outputs, we propose a novel weakly supervised contrastive loss for medical report generation. 
Experimental results demonstrate that our method benefits from contrasting target reports with incorrect but semantically-close ones.
It outperforms previous work on both clinical correctness and text generation metrics for two public benchmarks.

\end{abstract}

\section{Introduction}
Automated radiology report generation aims at generating informative text from radiologic image studies. 
It could potentially improve radiology reporting and alleviate the workload of radiologists.
Recently, following the success of deep learning in conditional text generation tasks such as 
image captioning~\citep{vinyals2015show,xu2015show}, many methods have been proposed for this task~\citep{jing2017automatic,li2018hybrid, liu2019clinically, jing2020show, ni2020learning, chen2020generating}.

Unlike conventional image captioning benchmarks (e.g.~MS-COCO~\citep{lin2014microsoft}) where referenced captions are usually short, radiology reports are much longer with multiple sentences, which pose higher requirements for information selection, relation extraction, and content ordering. 
To generate informative text from a radiology image study, a caption model is required to understand the content, identify abnormal positions in an image and organize the wording to describe findings in images. However, the standard approach of training an encoder-decoder model with teacher forcing and cross-entropy loss often leads to 
text generation outputs with high frequency tokens or sentences appearing too often~\citep{ranzato2015sequence, holtzman2019curious}. 
This problem could be worse for chest X-ray report generation, since the task has a relatively narrow text distribution with domain-specific terminology and descriptions for medical images, and models often struggle to generate long and diverse reports ~\citep{harzig2019addressing, boag2020baselines}.

To tackle these challenging issues, we propose to introduce contrastive learning into chest X-ray report generation. 
However, simply using random non-target sequences as negative examples in a contrastive framework is suboptimal~\citep{lee2020contrastive}, as random samples are usually easy to distinguish from the correct ones. 
Hence, we further introduce a weakly supervised contrastive loss that assigns more weights to reports that are semantically close to the target.
By exposing the model to these ``hard'' negative examples during training, it could learn robust representations which capture the essence of a medical image and generate high-quality reports with improved 
performance on clinical correctness for unseen images.

Overall, our contributions are three-fold: 
\begin{itemize}
    \vspace{-0.5ex}
    \item We propose a novel objective for training a chest X-ray report generation model with a contrastive term, which contrasts target reports with incorrect ones during training.
    \vspace{-0.5ex}
    \item We develop a 
    weakly supervised method to identify ``hard'' negative samples and assign them with higher weights in our contrastive loss to further encourage diversity.  
    \vspace{-0.5ex}
    \item We conduct comprehensive experiments to show the effectiveness of our method, which outperforms existing methods on both clinical correctness and text generation metrics.
\end{itemize}

\begin{figure*}[t]
\vspace{-3ex}
\centering
\includegraphics[width=0.85\textwidth]{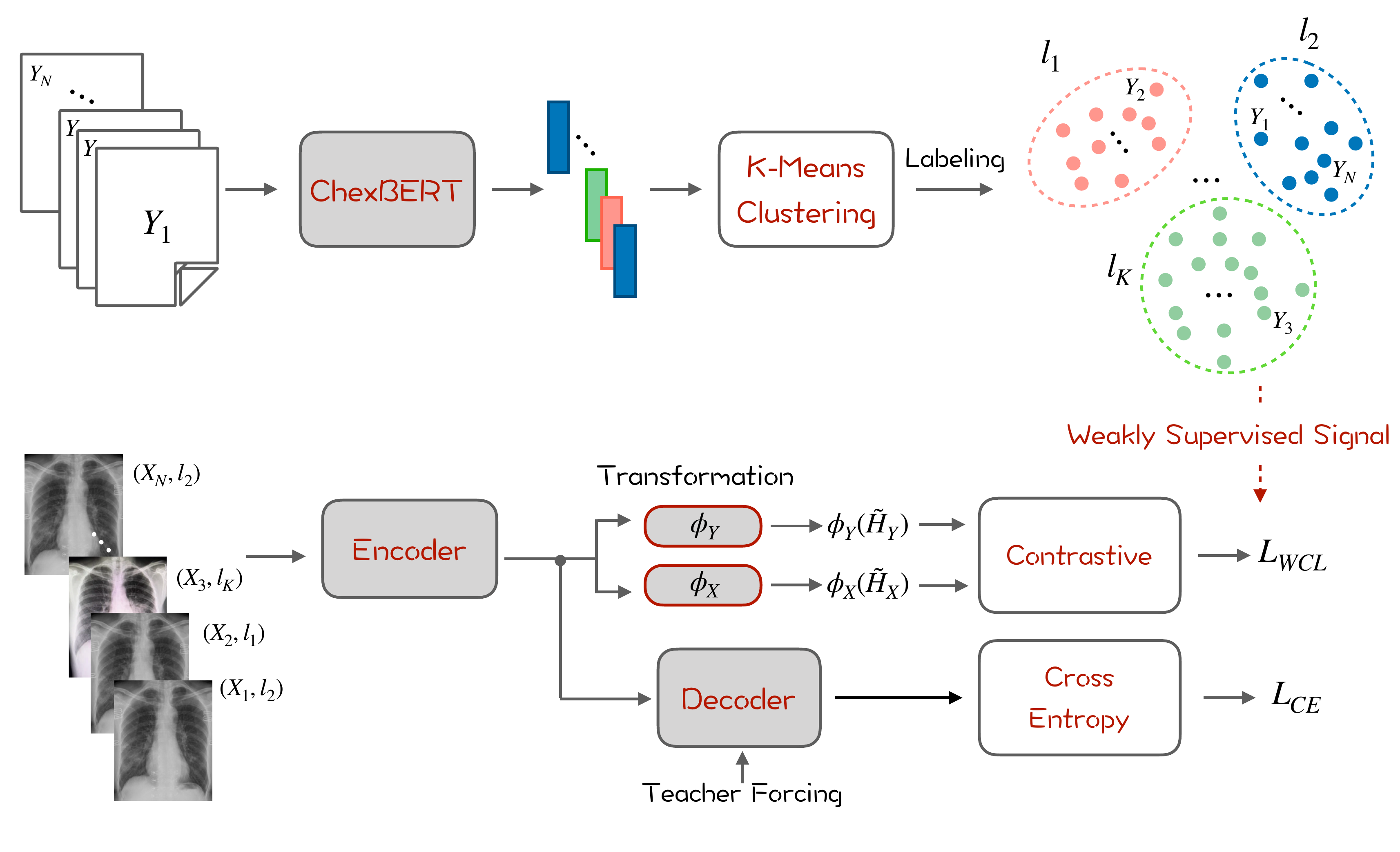}
\vspace{-3ex}
\caption{Illustration of our weakly supervised contrastive learning framework. We use a task-specific BERT model to label the reports, guiding the contrastive learning process during training.}
\label{fig:model}
\end{figure*}

\section{Related Work}
\paragraph{Medical Report Generation}
Medical report generation, which aims to automatically generate descriptions for clinical radiographs (e.g.~chest X-rays), has drawn attention in both the machine learning and medical communities.
Many methods have been proposed to solve this task.
For example, \citet{jing2017automatic} proposed a co-attention hierarchical CNN-RNN model. 
\citet{li2018hybrid, li2019knowledge} used hybrid retrieval-generation models to complement generation. Most recently, \citet{ni2020learning} proposed to learn visual semantic embeddings for cross-modal retrieval in a contrastive setting, and showed improvements in identifying abnormal findings. However, this is hard to scale or generalize since we need to build template abnormal sentences for new datasets.
\citet{chen2020generating} leveraged a memory-augmented transformer model to improve the ability to generate long and coherent text, but didn't address the issue of generating dominant normal findings specifically.
Our work proposes to incorporate contrastive learning into training a generation-based model, which benefits from the contrastive loss to encourage diversity and is also easy to scale compared to retrieval-based methods.

\paragraph{Contrastive Learning}
Contrastive learning~\citep{gutmann2010noise, oord2018representation} has been widely used in many fields of machine learning. 
The goal is to
learn a representation by contrasting positive and negative pairs. 
Recent work showed that contrastive learning can boost the performance of self-supervised and semi-supervised learning in computer vision tasks~\citep{he2020momentum, chen2020simple, khosla2020supervised}. 
In natural language processing, contrastive learning has been investigated for several tasks, including language modeling~\citep{huang2018large}, unsupervised word alignment~\citep{liu2015contrastive} and machine translation~\citep{yang2019reducing, lee2020contrastive}. In this work, we are interested in applying contrastive learning to chest X-ray report generation in a multi-modality setting. Different from previous work in applying contrastive learning to image captioning~\citep{dai2017contrastive}, we leverage a recent 
contrastive formulation inspired by~\citep{chen2020simple} which transforms representations into latent spaces, and propose a novel learning objective for the medical report generation task specifically.

\section{Method}

\paragraph{Generating Reports with Transformer}
We leverage a memory-driven transformer proposed in~\citep{chen2020generating} as our backbone model, which uses a memory module to record key information when generating long texts. 

Given a chest X-ray image $I$, its visual features $X$ are extracted by pre-trained convolutional neural networks (e.g. ResNet~\citep{he2016deep}).  
Then we use the standard encoder in transformer to obtain hidden visual features $H_X$. 
The decoding process at each time step $t$ can be formalized as 
\begin{equation}
    \hat{y}_t = \mathit{Decoder}(H_X, y_1, \ldots, y_{t-1}).
\end{equation}  
We use a cross-entropy (CE) loss to maximize the conditional log likelihood $\log p_{\theta}(Y|X)$ for a given $N$ observations ${(X^{(i)}, Y^{(i)}}_{i=1})^N$ as follows:
\begin{equation}
    \mathcal{L}_{CE} = \sum_{i=1}^N  \log p_\theta (Y^{(i)} | X^{(i)}).
\end{equation}

\paragraph{Labeling Reports with Finetuned BERT}
As shown in \cref{fig:model}, we first extract the embeddings of each report from ChexBERT~\citep{smit2020chexbert}, a BERT model pretrained with biomedical text and finetuned for chest X-ray report labeling.  
We use the [CLS] embedding of BERT to represent report-level features. We then apply K-Means to cluster the reports into $K$ groups. 
After clustering, each report $Y$ is assigned with a corresponding cluster label $l$, where reports in 
the same cluster are 
considered to be
semantically close to each other.

\paragraph{Weakly supervised Contrastive Learning}
To regularize the training process, we propose a weakly supervised contrastive loss (WCL). 
We first project the hidden representations of the image and the target sequence into a latent space:
\begin{equation}
    z_x = \phi_x (\tilde{H}_X), z_y = \phi_y (\tilde{H}_Y),
\end{equation}
where $\tilde{H}_X$ and $\tilde{H}_Y$ are the average pooling of the hidden states $H_X$ and $H_Y$ from the transformer, $\phi_x$ and $\phi_y$ are two fully connected layers with ReLU activation~\citep{nair2010rectified}.
We then maximize the similarity between the pair of source image and target sequence, while minimizing the similarity between the negative pairs as follows:
\begin{equation}
    \mathcal{L}_{\mathit{WCL}} = \sum_{i=1}^N  \log \frac{\exp(s_{i,i})}{\sum\limits_{l_i \neq l_j} \exp (s_{i,j}) + \alpha \sum\limits_{l_i=l_j} \exp (s_{i,j})}, 
    \label{eqn:contra}
\end{equation}
where $s_{i,j} = \mathit{sim}({z_x}^{(i)}, {z_y}^{(j)})/\tau$, $\mathit{sim}$ is the cosine similarity between two vectors, $\tau$ is the temperature parameter, $\alpha$ is a hyperparameter that weighs the importance of negative samples that are semantically close to the target sequence, i.e., with the same cluster label $l_i=l_j$ in \cref{eqn:contra}. Empirically, we find that these samples are  ``hard'' negative samples and the model would perform better by assigning more weights to distinguish these samples.

Overall, the model is optimized with a mixture of cross-entropy loss and weakly supervised contrastive loss:
\begin{equation}
    \begin{split}
    \mathcal{L}_{loss} = (1-\lambda) \mathcal{L}_{\mathit{CE}} +  \lambda \mathcal{L}_{\mathit{WCL}},
    \end{split}
\end{equation}
where $\lambda$ is a hyperparameter that weighs the two losses.

\section{Experiments}

\begin{table*}[t]
\small
\centering
\setlength{\tabcolsep}{8pt}
\begin{tabular}{l c rrrr rrrrr}
\toprule
\multirow{2}{*}{\textbf{Dataset}} & \multirow{2}{*}{\textbf{Model}} & \multicolumn{4}{c}{\textbf{NLG metrics}} & \multicolumn{3}{c}{\textbf{CE metrics}}\\
\cmidrule(lr){3-6} \cmidrule(lr){7-9}
 &  & BLEU-1 & BLEU-4 & METEOR & ROUGE-L & Precision  & Recall  & F-1 \\
\toprule
\multirow{5}{*}{\textbf{MIMIC-ABN}}
& \textit{ST} & 14.9 & 3.3 & 7.2 & 17.4 & 20.3 & 22.2 & 21.2 \\
& \textit{HCR}  & 8.4 & 1.9 & 5.9 & 14.9 & 26.1 & 15.7 & 19.6 \\
& \textit{CVSE} & 19.2 & 3.6 & 7.7 & 15.3 & 31.7 & 22.4 & 26.2  \\
& \textit{MDT} & 24.6 & 6.6  & 9.7 & 23.0  & \textbf{34.0} & 29.1 & 29.4 \\
& \textit{MDT+WCL} & \textbf{25.6} & \textbf{6.7} & \textbf{10.0} & \textbf{24.1} & 33.2 & \textbf{30.9}  & \textbf{30.0} \\
\midrule 
\multirow{4}{*}{\textbf{MIMIC-CXR}}
& \textit{ST}  & 29.9 & 8.4 & 12.4 & 26.3 & 24.9 & 20.3 & 20.4\\
& \textit{TopDown} & 31.7 & 9.2 & 12.8 & 26.7 & 32.0 & 23.1 & 23.8 \\
& \textit{MDT} & 35.3 & 10.3 & 14.2 & \textbf{27.7} & 33.3 & 27.3 & 27.6 \\
& \textit{MDT+WCL} & \textbf{37.3} & \textbf{10.7} & \textbf{14.4}  & 27.4 & \textbf{38.5} & \textbf{27.4} & \textbf{29.4} \\
\bottomrule
\end{tabular} 
\caption{The performance of all baselines and our full model on the test sets of MIMIC-ABN and MIMIC-CXR datasets with respect to natural language generation (NLG) and clinical efficacy (CE) metrics. Results are reported in percentage~(\%). \textit{ST} is CNN+LSTM with attention~\citep{xu2015show}. \textit{HCR}~\citep{jing2017automatic} is a hierachical CNN-RNN model. \textit{CVSE}~\citep{ni2020learning} is a cross-modal retrieval model. \textit{TopDown}~\citep{anderson2018bottom} is a widely-used image captioning model. \textit{MDT} is a memory-driven transformer proposed in \citep{chen2020generating}.
}
\label{tab:main}
\end{table*}

\subsection{Experimental Setup}
\paragraph{Datasets} 
We conduct experiments on two datasets: (1) MIMIC-ABN, which was proposed in~\citep{ni2020learning} and contains a subset of images of MIMIC-CXR with abnormal sentences only, with 26,946/3,801/7,804 reports for train/val/test sets. (2) MIMIC-CXR~\citep{johnson2019mimic}, the largest radiology dataset to date that consists of 222,758/1,808/3,269 reports for train/val/test sets.

\paragraph{Evaluation Metrics}
Performance is first evaluated on three automatic metrics: BLEU~\citep{papineni2002bleu}, ROUGE-L~\citep{lin2004rouge}, and METEOR~\citep{denkowski2011meteor}.

We then use the CheXpert labeler to evaluate the clinical accuracy of the abnormal findings reported by each model, which is a state-of-the-art rule-based chest X-ray report labeling system~\citep{irvin2019chexpert}. Given sentences of abnormal findings, CheXpert will give a positive and negative label for 14 diseases. We then calculate the Precision, Recall and Accuracy for each disease based on the labels obtained from each model’s output and from the ground-truth reports.

\paragraph{Implementation Details}
We use Adam as the optimizer with 
initial learning 
rates of 5e-5 for the visual extractor 
and 1e-4 for other parameters.
The learning rate is decayed by a factor of 0.8 per epoch.
We set the number of labels $K$ to 13 and 14 for MIMIC-ABN and MIMIC-CXR, the weighting parameters $\lambda$ and $\alpha$ to 0.2 and 2 respectively.
We conduct a grid-based hyperparameter search for weighting factor $\lambda \in \{0.1, 0.2, 0.3, 0.4, 0.5\}$ and temperature $\tau \in \{0.1, 1, 10\}$ by evaluating the models on the validation sets of the two datasets. 

Words with frequency less than 3 and 10 are discarded for MIMIC-ABN and MIMIC-CXR respectively.
The maximum sequence lengths are set to 64 and 100 for MIMIC-ABN and MIMIC-CXR. 
The projection heads consists of two convolutional layers with ReLU activation, and the latent dimensions are set to 256 for both visual and text projection layers.
The hidden sizes of both the encoder and decoder are 512 with 8 heads and 3 layers. The batch size for training the two datasets is 128. We report the results using models with the highest BLEU-4 on the validation set. We use a beam size of 3 for generation to balance the generation effectiveness and efficiency.

\subsection{Performance comparison}
We compare our approach to other methods on two datasets. As shown in \cref{tab:main}, first, our method (\textit{MDT+WCL}) outperforms previous retrieval (\textit{CVSE}) and generation based models (\textit{MDT}) on most text generation metrics.
Second, our contrastive loss significantly improves clinical efficacy metrics, demonstrating its capability to accurately report abnormal findings.
Finally, the relative difference between \textit{MDT} and \textit{MDT+WCL} is higher on 
MIMIC-CXR, 
which contains
a larger training set for learning robust representations. 

\subsection{Analysis}
\paragraph{Different Contrastive Losses}
\begin{table}[t]
\small
\centering
\setlength{\tabcolsep}{2pt}
\begin{tabular}{l rrrr}
\toprule
  \multirow{2}{*}{\textbf{Method}} & \multicolumn{4}{c}{\textbf{Validation}}\\
\cmidrule(lr){2-5}
 & BLEU-1 & BLEU-4 & METEOR & ROUGE-L \\
\toprule
\textit{baseline} & 25.0 & 6.6 & 9.7 & 23.3 \\
\textit{adversarial} & 25.1 & 6.6 & 9.8 & 23.2 \\
\textit{excluding} & 24.8 & 6.6 & 9.8 & 23.1 \\
\midrule 
\textit{Ours} & \textbf{25.4} & \textbf{6.8} & \textbf{10.0} & \textbf{24.0} \\
\bottomrule
\end{tabular}
\caption{Ablation study on the MIMIC-ABN dataset. \textit{baseline} is the vanilla contrastive loss ($\alpha=1$) introduced in~\citep{chen2020simple}; \textit{excluding} excludes negative samples in the same cluster ($\alpha=0$); \textit{adversarial} is proposed in~\citep{lee2020contrastive} where they construct extra negative samples via adversarial attack. Results are averaged over three runs.
}
\label{tab:ablation}
\end{table}

We compare our contrastive formulation with other variants in \cref{tab:ablation}. 
Our approach achieves the best performance over all baselines on different metrics. 
Both \textit{adversarial} and \textit{excluding} have similar performance compared to the vanilla contrastive framework~\citep{chen2020simple}, not showing improvements for medical report generation.
On the other hand, we identify the reports in the same cluster as ``hard'' negative samples and assign more weights on them during training, guiding the model to better distinguish reference reports from inaccurate ones.

\begin{figure}[t]
  \centering
  \vspace{-2ex}
  \includegraphics[width=\linewidth]{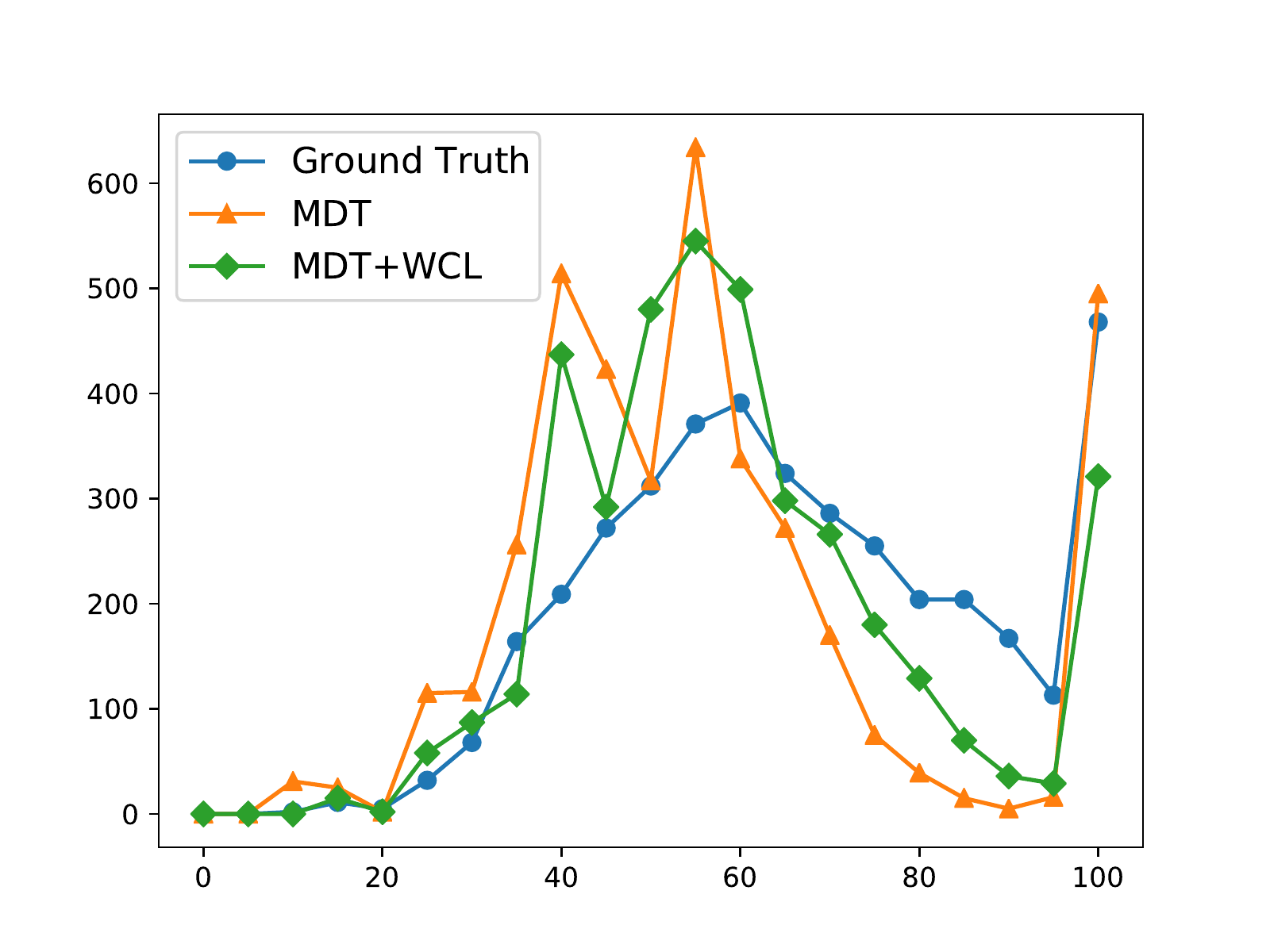}
  \caption{The length distributions of the generated reports on the MIMIC-CXR test set.}
  \label{fig:length}
\end{figure}

\paragraph{Length Distributions}
To further evaluate the generation quality of our method in addition to NLG and CE metrics, we compare the length of generated reports to the ground truth. To do this, we categorize all reports generated on the MIMIC-CXR test set into 20 groups (within the range of [0, 100] with an interval of 5). As shown in \cref{fig:length}, the baseline model \textit{MDT} has a sharper distribution, while adding our weakly supervised contrastive loss leads to a length distribution which is smoother and closer to the ground truth, demonstrating its effectiveness to generate more diverse and accurate reports, and generalize on unseen images.

\paragraph{Case Study}
\begin{table*}[t]
\small
\centering
\setlength{\tabcolsep}{2pt}
\begin{tabular}{p{0.9\textwidth} }
\toprule
\midrule 
\textbf{GT}: streaky left basilar opacity likely reflects atelectasis. minimal left basilar atelectasis. history. with history of myocardial infarction presenting with epigastric pain. \\
\textbf{WCL}: linear left basilar opacity is likely atelectasis versus scarring. m with chest pain.\\
\midrule 
\textbf{GT}: the lungs are hyperexpanded but clear consolidation effusion or pneumothorax. increased lucency at the left lung apex and linear markings on the lateral raises the possibility apical bullous disease. hyperexpansion without acute cardiopulmonary process. \\
\textbf{WCL}: the lungs are hyperinflated but clear of consolidation or effusion. hyperinflation without acute cardiopulmonary process. \\
\midrule 
\textbf{GT}: there is bibasilar atelectasis. a linq cardiac monitoring device projects over the subcutaneous tissue of the left lower chest. f with shortness of breath. evaluate for pneumonia. \\
\textbf{WCL}: the lungs are hyperinflated with flattening of the diaphragms suggesting chronic obstructive pulmonary disease. there is mild bibasilar atelectasis. \\
\midrule 
\bottomrule
\end{tabular}
\caption{Generated samples from MIMIC-ABN dataset.
}
\label{tab:abn-samples}
\end{table*}

\begin{table*}[t]
\small
\centering
\setlength{\tabcolsep}{2pt}
\begin{tabular}{p{0.9\textwidth} }
\toprule
\midrule 
\textbf{GT}: there is moderate amount of right-sided subcutaneous emphysema which is similar in appearance compared to prior. right-sided chest tube is again visualized. there is no increase in the pneumothorax. bilateral parenchymal opacities are again visualized and not significantly changed. the tracheostomy tube is in standard location. right subclavian line tip is in the mid svc.
\\
\textbf{WCL}: tracheostomy tube tip is in unchanged position. right-sided port-a-cath tip terminates in the low svc. left-sided port-a-cath tip terminates in the proximal right atrium unchanged. heart size is normal. mediastinal and hilar contours are similar. innumerable bilateral pulmonary nodules are re- demonstrated better assessed on the previous ct. small right pleural effusion appears slightly increased compared to the prior exam. small left pleural effusion is similar. no new focal consolidation or pneumothorax is present. there are no acute osseous abnormalities.\\
\midrule 
\textbf{GT}: the lungs are mildly hyperinflated as evidenced by flattening of the diaphragms on the lateral view. diffuse interstitial markings compatible with known chronic interstitial lung disease are unchanged. there is no pleural effusion or evidence of pulmonary edema. there is no focal airspace consolidation worrisome for pneumonia. mild to moderate cardiomegaly is unchanged. the mediastinal and hilar contours are unremarkable. a coronary artery stent is noted. there is. levoscoliosis of the thoracic spine . \\
\textbf{WCL}: lung volumes are low. heart size is mildly enlarged. the aorta is tortuous and diffusely calcified. crowding of bronchovascular structures is present without overt pulmonary edema. patchy opacities in the lung bases likely reflect areas of atelectasis. no focal consolidation pleural effusion or pneumothorax is present. there are no acute osseous abnormalities.\\
\midrule 
\bottomrule
\end{tabular}
\caption{Generated samples from MIMIC-CXR dataset.
}
\label{tab:cxr-samples}
\end{table*}

We present examples of generated reports and their corresponding ground truth from MIMIC-ABN and MIMIC-CXR. As shown in \cref{tab:abn-samples} and \cref{tab:cxr-samples}, our method (WCL) is able to generate similar contents which are aligned with the ground truth (GT) written by radiologists. For example, abnormal findings in specific positions (e.g., ``low lung volumes'' and ``enlarged heart size'') are reported, and potential diseases are also noted (e.g., ``atelectasis''). 

\section{Conclusion}
In this paper, we present a weakly supervised contrastive learning framework for generating chest X-ray reports. 
Our contrastive loss could lead to better results on both clinical correctness and text generation metrics than previous methods.
We also show that exposing the model to semantically-close negative samples improves generation performance.
In the future, we will
extend our method to other medical image datasets other than chest X-rays.

\section*{Acknowledgments}
This work was partially supported by the Office of the Assistant Secretary of Defense for Health Affairs through the AIMM Research Program endorsed by the Department of Defense under Award No. W81XWH-20-1-0693 and NSF Award \#1750063. Opinions, interpretations, conclusions and recommendations are those of the author and are not necessarily endorsed by the Department of Defense and the National Science Foundation.

\bibliography{anthology, emnlp2021}
\bibliographystyle{acl_natbib}




\end{document}